\documentclass[10pt,twocolumn,letterpaper]{article}

\usepackage[pagenumbers]{cvpr} 

\usepackage{graphicx}
\usepackage{amsmath}
\usepackage{amssymb}
\usepackage{booktabs}
\usepackage{float}
\usepackage{times}
\usepackage{pifont}
\usepackage{enumitem}
\usepackage{multirow}
\usepackage[table]{xcolor}
\newcommand{\cmark}{\ding{51}}
\newcommand{\framework}[1]{Pseudo-Q}
\definecolor{myblue}{rgb}{0.27, 0.80, 1.0}
\definecolor{mygreen}{rgb}{0.6, 1.0, 0.6}
\definecolor{myred}{rgb}{1.0, 0.2, 0.2}
\usepackage[pagebackref,breaklinks,colorlinks]{hyperref}

\def\cvprPaperID{3407} 
\def\confName{CVPR}
\def\confYear{2022}

\begin{document}

\title{Pseudo-Q: Generating Pseudo Language Queries for Visual Grounding}

\author{%
  Haojun Jiang$^{1}$\thanks{Equal contribution.}\ \ \
  Yuanze Lin$^{3}$\footnotemark[1] \thanks{This work was done during an internship at Tsinghua.}\ \ \
  Dongchen Han$^{1}$\ \ \
  Shiji Song$^{1}$\ \ \
  Gao Huang$^{1,2}$\thanks{Corresponding author.} \\
    $^{1}$Tsinghua University, BNRist \ \ \
    $^{2}$BAAI \ \ \
    $^{3}$University of Washington \\
  \texttt{\small \{jhj20, hdc19\}@mails.tsinghua.edu.cn}, \texttt{\small yuanze@uw.edu},\\
  \texttt{\small\{shijis, gaohuang\}@tsinghua.edu.cn}
}

\maketitle

\begin{abstract}
Visual grounding, i.e., localizing objects in images according to natural language queries, is an important topic in visual language understanding. The most effective approaches for this task are based on deep learning, which generally require expensive manually labeled image-query or patch-query pairs. To eliminate the heavy dependence on human annotations, we present a novel method, named \textit{\framework{}}, to automatically generate pseudo language queries for supervised training. Our method leverages an off-the-shelf object detector to identify visual objects from unlabeled images, and then language queries for these objects are obtained in an unsupervised fashion with a pseudo-query generation module. Then, we design a task-related query prompt module to specifically tailor generated pseudo language queries for visual grounding tasks. Further, in order to fully capture the contextual relationships between images and language queries, we develop a visual-language model equipped with multi-level cross-modality attention mechanism. Extensive experimental results demonstrate that our method has two notable benefits: (1) it can reduce human annotation costs significantly, \textit{e.g.}, \textit{31$\%$} on RefCOCO~\cite{yu2016modeling} without degrading original model's performance under the fully supervised setting, and (2) without bells and whistles, it achieves superior or comparable performance compared to state-of-the-art weakly-supervised visual grounding methods on all the five datasets we have experimented. Code is available at \url{https://github.com/LeapLabTHU/Pseudo-Q}.

\end{abstract}

\section{Introduction}
Visual grounding (VG) task \cite{mao2016generation,yu2016modeling,hu2016natural,deng2021transvg} has achieved great progress in recent years, with the advances in both computer vision \cite{huang2017densely,he2016deep,ren2015faster,dosovitskiy2020image,yang2021condensenet,han2021spatially,huang2022glance,wang2021adaptive,wang2021adafocus} and natural language processing \cite{mikolov2013distributed,sutskever2014sequence, vaswani2017attention,devlin2018bert,brown2020language}. It aims to localize the objects referred by natural language queries, which is essential for various vision-language tasks, \textit{e.g.}, visual question answering \cite{antol2015vqa} and visual commonsense reasoning \cite{zellers2019recognition}.

\begin{figure}[t]
\centering
\includegraphics[scale=0.38]{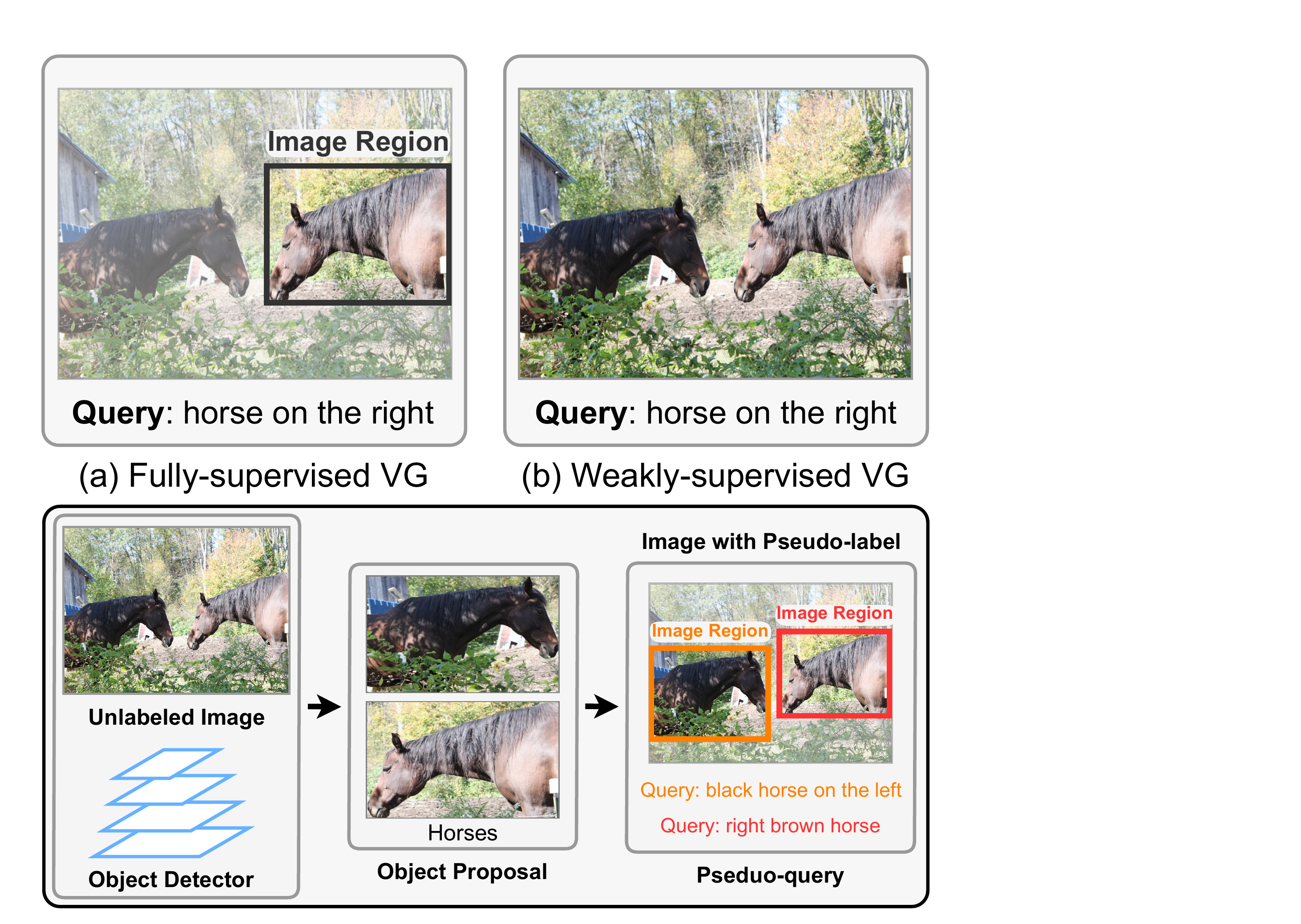}{\centering}
\caption{\textbf{Comparison with fully and weakly-supervised visual grounding method}. (a) Fully-supervised VG utilizes image region-query pairs as supervision signals. (b) Weakly-supervised VG adopts only language queries. (c) Our \textbf{\framework{}} method is free of any task-related annotations.}
\label{fig:1}
\centering
\vspace{-10pt}
\end{figure}

Most existing visual grounding methods can be categorized into two types: fully-supervised \cite{chen2018real,liao2020real,hong2019learning,hu2017modeling,liu2019learning,deng2021transvg} and weakly-supervised \cite{xiao2017weakly,chen2018knowledge,datta2019align2ground,liu2019adaptive,gupta2020contrastive,liu2021relation,wang2021improving,sun2021discriminative}. Although these two lines of works have made remarkable successes, they rely heavily on manually annotated datasets. However, obtaining a large quantity of manual annotations, especially natural language queries, is expensive and time-consuming. To annotate queries, humans need to firstly recognize the visual objects and identify their attributes, and then determine diverse relationships between them on a case-by-case basis, such as spatial (\textit{e.g.}, \emph{left} and \emph{right}), preposition (\textit{e.g.}, \emph{in} and \emph{with}), action (\textit{e.g.}, \emph{throwing something}), and comparative (\textit{e.g.}, \emph{smaller} and \emph{bigger}). Among them, \textit{spatial} relation is the most frequently queried one. 

To reduce the burden of human annotation, we propose a pseudo language query based approach (\textbf{\framework{}}) for visual grounding. Our inspiration comes from previous works \cite{feng2019unsupervised, laina2019towards} that address the high annotation cost issue in image captioning task, by leveraging an unlabelled image set, a sentence corpus, and an off-the-shelf object detector. However, the visual grounding task is more complicated and challenging, as it involves the modelling of relations between objects.

To accurately ground objects by language queries, it's fundamental to recognize their categories, attributes, and relationships. Thus, when it comes to generating pseudo region-query pairs for an unlabelled image set, we need to focus on three key components: (1) salient \textbf{objects (nouns)} which are most likely to be queried, (2) intrinsic \textbf{attributes} possessed by the queried objects, and (3) the important \textbf{spatial relationships} between the objects. Motivated by \cite{feng2019unsupervised,nam2021zero}, we leverage an off-the-shelf object detector~\cite{anderson2018bottom} to locate the most notable candidates with high confidence, and an attribute classifier~\cite{anderson2018bottom} to recognize common attributes. However, these detectors are unable to distinguish the spatial relations between objects. Thus, we present a heuristic algorithm to determine the \emph{spatial relationships} between the objects of the same class by comparing their areas and relative coordinates. With these three essential components, pseudo-queries with respect to spatial relations between objects can be generated.

To further improve the performance of our method, we also propose a query prompt module  which attentively tailors generated pseudo queries into task-related query templates for visual grounding. For the visual-language model, we put forward a multi-level cross-modality attention mechanism in the fusion module to encourage a deeper fusion between visual and language features.

Extensive experiments have verified the effectiveness of our method. First, in fully supervised manner, it can reduce human annotation costs by \textit{31$\%$} without sacrificing original model's performance on RefCOCO\cite{yu2016modeling}. Second, without bells and whistles, it can obtain superior or comparable performance even compared with state-of-the-art weakly-supervised visual grounding methods on five datasets, including RefCOCO \cite{yu2016modeling}, RefCOCO+ \cite{yu2016modeling}, RefCOCOg \cite{mao2016generation}, ReferItGame \cite{kazemzadeh2014referitgame} and Flickr30K Entities \cite{plummer2015flickr30k}.

In summary, this paper makes three-fold contributions:
\vspace{-3pt}
\begin{enumerate}
    \setlength\itemsep{-2pt}
    \item [(1)]
    We introduce the first pseudo-query based visual grounding method that deals with the most dominant spatial relationships among objects.
    \item [(2)]
    We propose a query prompt module to specifically tailor pseudo-queries for visual grounding task, and a visual-language model equipped with multi-level cross-modality attention is put forward to fully capture the contextual relationships of different modalities.
    \item [(3)]
    Extensive experiments demonstrate that our approach can not only dramatically reduce the manual labelling costs without performance sacrifice under the fully supervised condition, but also surpass or achieve comparable performance with state-of-the-art weakly-supervised visual grounding methods.
\end{enumerate}

\section{Related Work}
\subsection{Natural Language Visual Grounding} 
Visual grounding is a crucial component in vision and language, and it serves as the fundamental of other tasks, such as VQA. Recent visual grounding methods can be summarized into three categories: fully-supervised~\cite{chen2018real, liao2020real, hong2019learning, hu2017modeling, liu2019learning, deng2021transvg}, weakly-supervised~\cite{xiao2017weakly, chen2018knowledge, datta2019align2ground, liu2019adaptive, gupta2020contrastive, liu2021relation, wang2021improving, sun2021discriminative}, and unsupervised~\cite{yeh2018unsupervised, wang2019phrase}. Fully-supervised methods rely heavily on the manual labeled patch-query pairs. Unfortunately, obtaining such sophisticated annotations is expensive and time-consuming. Consequently, weakly-supervised approaches attempt to alleviate the issue by utilizing only image-query pairs. These methods\cite{chen2018knowledge, liu2021relation} usually leverage a mature object detector to compensate the missing bounding box labels for training. However, annotating the language queries for salient objects in an image is the most laborious part. Thus, unsupervised methods~\cite{yeh2018unsupervised, wang2019phrase} attempt to train a model or directly detect queried objects without any annotations. Our work is also an unsupervised method. However, unlike previous methods, we present a novel method, named \framework{}, to automatically generate pseudo-queries for supervised learning.

\begin{figure*}[t]
\centering
\includegraphics[scale=0.305]{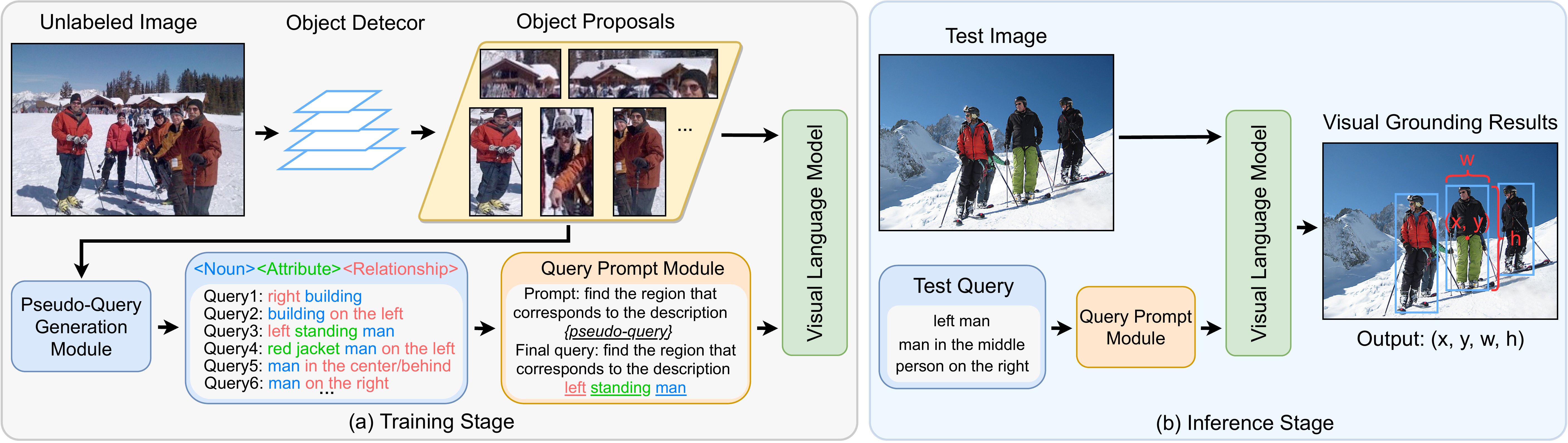}
\caption{\textbf{Overview of our Pseudo-Q method. Better view in color and zoom in.} The proposed approach consists of a pseudo-query generation module, a query prompt module, and a visual-language model. (a) During the training stage, pseudo image region-query pairs are generated to train visual language model. (b) During the inference stage, the test query is filled into the prompt template, and the target object is located by the trained model. }
\label{fig:frameowrk}
\centering
\vspace{-10pt}
\end{figure*}

\subsection{Vision-Language Transformer}
Transformer \cite{vaswani2017attention} has been firstly proposed to address natural language processing (NLP) tasks. ViT \cite{dosovitskiy2020image} makes the first attempt to apply a transformer for image classification task \cite{deng2009imagenet}. Motivated by the success of ViT, DETR \cite{carion2020end} and Segmenter \cite{strudel2021segmenter} further extend the transformer for object detection and segmentation tasks respectively. 

There are also many efforts \cite{chen2020uniter, li2020oscar, lu2019vilbert, deng2021transvg, radford2021learning, tan2019lxmert}, which try to handle visual-language tasks by transformer. TransVG \cite{deng2021transvg} proposes a novel framework with transformer structure for visual grounding task. CLIP \cite{radford2021learning} and UNITER \cite{chen2020uniter} leverage transformers for jointly learning text and image representations. LXMERT \cite{tan2019lxmert} establishes a large-scale transformer to learn cross-modality representation. In this work, we propose a novel multi-level cross-modality attention on the top of the TransVG for cross-modality learning. 

\subsection{Visual Recognition without Annotation}
There have been several works~\cite{chen2021elaborative, zhang2020zstad, teney2016zero, nam2021zero, demirel2017attributes2classname, dixit2019semantic, jain2015objects2action, bansal2018zero, gu2021zero} for zero-shot visual tasks. Zero-shot object detection tasks \cite{bansal2018zero, gu2021zero} are designed for detecting unseen object classes whose labels are missing. While zero-shot action recognition task \cite{ demirel2017attributes2classname, dixit2019semantic, jain2015objects2action} recognizes pre-defined action categories without using action labels. Our work's emphasis lies in locating object regions without using any task-related annotations, \textit{e.g.}, image regions and queries. 

As for zero-shot visual grounding, the pioneering work ZSGNet \cite{sadhu2019zero} focuses on query phrases which may contain unseen nouns or object categories. It consists of a language module to encode query features, a visual module to extract image features, and an anchor generator to produce anchors. However, note that the focus of our work is different from ZSGNet, which is proposed for recognizing unseen classes, In addition, ZSGNet utilizes manual annotations while we do not rely on any task-related labels.

\begin{figure*}[t]
\centering
\includegraphics[scale=0.29]{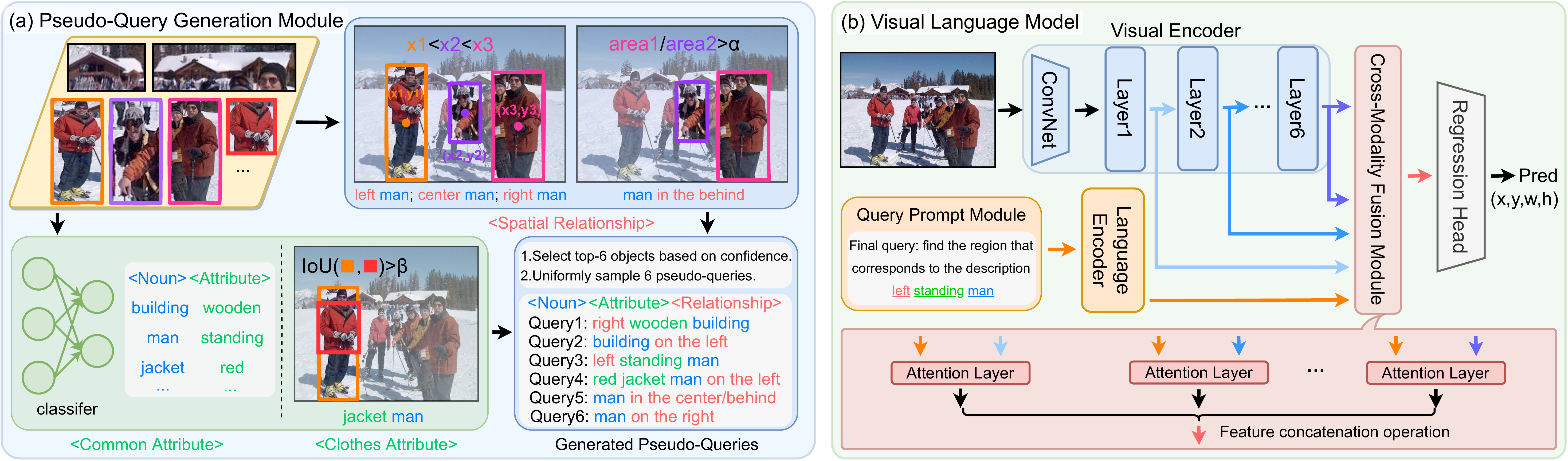}
\caption{(a) The \textbf{pseudo-query generation module} produces spatial relationships and attributes for corresponding objects. (b) The \textbf{visual-language model} consists of a visual encoder, a language encoder, and a cross-modality fusion module.}
\label{fig:details_of_framework}
\centering
\vspace{-5pt}
\end{figure*}
\vspace{5pt}
\section{Method}
In this section, we explain our \framework{} method in detail. In Sec.~\ref{framework}, we introduce the overview of Pseudo-Q. In Sec.~\ref{pseudo}, we elaborate the pseudo-query generation module. In Sec.~\ref{prompt}, the details of the task-related query prompt module are shown. Finally, we illustrate the mechanism of our multi-level cross-modality attention in Sec.~\ref{cross}.

\subsection{Overview}
\label{framework}
Previous visual grounding methods rely on expensive human annotations, \textit{i.e.}, image region-query pairs for fully-supervised approaches \cite{deng2021transvg, hong2019learning, liu2019learning} or image-query pairs for weakly-supervised approaches \cite{sun2021discriminative, liu2019knowledge, liu2019adaptive}. We firstly propose a pseudo language query based method without using any task-related annotations at training.

Specifically, the \framework{} approach consists of three components, including: (1) pseudo-query generation module, (2) query prompt module, and (3) visual-language model. The illustration of \framework{} is shown in Figure~\ref{fig:frameowrk}. Taking an unlabeled image as an explanation, the detector can produce several object proposals. Then, these proposals are fed into pseudo-query generation module, which can automatically generate \textit{nouns}, \textit{attributes}, and \textit{relationships} for these proposals. Together with these elements, we can easily create pseudo language queries. 

Subsequently, the query prompt module refines created pseudo language queries for visual grounding task. Finally, we propose a visual-language model to fully capture the contextual relationship between the image regions and corresponding pseudo language queries. 


\subsection{Pseudo-Query Generation}
\label{pseudo}
In general, the first step for visual grounding is recognizing the categories of queried objects. However, such a simple grounding strategy leads to ambiguities in complex scenes, \textit{e.g.}, \emph{``a talk person on the left"} or \emph{``a short person on the right"}, without understanding their spatial relations or attributes. Thus, to accurately locate visual objects by language queries, a visual grounding model needs to understand queried objects' categories, attributes, and their relationships. Based on the above analysis, generating pseudo language queries for candidate objects involving three key components: \textbf{nouns}, \textbf{attributes} and \textbf{relationships}.

\textbf{Nouns.} Inspired by works\cite{nam2021zero, feng2019unsupervised, laina2019towards}, we adopt an off-the-shelf detector~\cite{anderson2018bottom} to obtain the object proposals. Unlike image classification task where each image contains only one major object, scenes in visual grounding task are more complex due to plenty of candidate objects. While it is natural to select the most salient objects as candidates, such a process requires intensive manual labor which is not available in our setup. Instead, we use detection confidence as a criterion. Concretely, the top-$N$ objects with highest confidence are kept as our proposals. Furthermore, we empirically discover that the detector will focus on a large quantity of tiny objects which are less likely to be queried. Thus, we propose to remove tiny objects before generating proposals.

\textbf{Attributes.} They are important semantic cues that help models understand scenes better. We investigate that, in existing datasets \cite{yu2016modeling, kazemzadeh2014referitgame, mao2016generation}, common attributes including color, size (\textit{tall}), material (\textit{wooden}) and human state of motion ($e.g.$ \textit{standing} and \textit{walking}), etc. Similar to obtaining the nouns, we take advantage of an off-the-shelf attribute classifier~\cite{anderson2018bottom} to predict the above common attributes of corresponding objects. In general, one object may have several attributes, such as ``\textit{a tall person is walking}", and it is ideal to recognize as many attributes as possible. However, limited by the capability of the model, we only keep the attribute with the highest confidence and exceeding a predefined threshold as the final proposal. Furthermore, clothes are also important attributes for a person which can be determined by calculating the IoU value between clothes and person, as shown in Figure~\ref{fig:details_of_framework}(a). 

\textbf{Relationships.} We observe that \textit{spatial relationship} is one of the most frequently used relations in most existing datasets \cite{yu2016modeling, mao2016generation} to distinguish different objects. In order to excavate latent spatial relationships, we propose a heuristic algorithm as shown in Figure~\ref{fig:details_of_framework}(a). 

In general, spatial relationship can be divided into three dimensions: horizontal (\textit{i.e.}, \emph{left}, \emph{middle}, and \emph{right}), vertical (\textit{i.e.}, \emph{top} and \emph{bottom}) and depth (\textit{i.e.}, \emph{front} and \emph{behind}). Note that each previously generated object proposal is represented by a set of coordinates which naturally embrace spatial information. We can obtain the horizontal and vertical spatial relationships by comparing the center coordinates of objects along with these two dimensions. Meanwhile, to increase the robustness of the algorithm, the numerical difference of two objects' coordinates in the same dimension is required to be greater than a pre-defined threshold. Finally, we can determine the spatial relations, such as \textit{left}, \textit{right} and \textit{center}, for different visual objects of the same category.

In the depth dimension, we assume that, for the same kind of object, the closer the object is to a camera the larger the object region. Concretely, we calculate the ratio of the area of the largest object region to the smallest object region and set a threshold to determine whether there is a \textit{front} and \textit{behind} relationship. If the ratio exceeds the threshold, we assign \textit{front} and \textit{behind} relationships to the largest and smallest objects respectively. 

\textbf{Pseudo-queries.} After obtaining three key elements, we can generate all possible pseudo-queries for an image following the templates in Appendix. Finally, we sample up to $M$ pseudo image region-query pairs if the number of candidates is greater than $M$, otherwise, we sample all.

\subsection{Query Prompt Module}
\label{prompt}
With the advances of pre-trained language models~\cite{devlin2018bert, brown2020language}, prompt engineering is proposed to better utilize their learned knowledge at pre-training stage. Inspired by the recent success of prompt engineering in visual-language tasks, \textit{e.g.}, image-language pre-training~\cite{radford2021learning}, we propose a query prompt module to excavate the hidden knowledge of pre-trained language model (Sec.~\ref{cross}) by refining generated pseudo language queries for visual grounding task. 

While the prompt templates proposed in CLIP~\cite{radford2021learning} works well for the image classification task, we empirically find that they are ineffective for the challenging visual grounding task. Consequently, in this work, we explore prompt templates exclusively for visual grounding. Our introduced query prompt module follows certain templates, \textit{e.g.}, \emph{``find the region that corresponds to the description \{pseudo-query\}"} or \emph{``which region does the text \{pseudo-query\} describe?"}. Such design is specifically tailored for visual grounding task, since the focus of this task lies on locating the regions of referred objects. 

\subsection{Visual-Language Model}
\label{cross}
Our visual-language model consists of a visual encoder, a language encoder and a  cross-modality fusion module to fuse information from two modalities. The designs of the visual encoder and the language encoder are following TransVG~\cite{deng2021transvg}. We elaborate them in Appendix.

\textbf{Cross-modality fusion module.} Previous method~\cite{deng2021transvg} naively utilizes final features of visual and language encoders to acquire cross-modality information. However, such a simple approach is suboptimal, since each level of visual feature possesses valuable semantic information~\cite{he2016deep, lin2017feature}. To be more specific, low-level features usually denote coarse information, \textit{e.g.}, shape and edge, while high-level features can represent finer information, \textit{e.g.}, intrinsic object properties. Thus, we further propose multi-level cross-modality attention (ML-CMA) to thoroughly fuse textual embedding with multi-level visual features. 

The mechanism of ML-CMA is shown in Figure~\ref{fig:details_of_framework}(b). Features of each visual transformer layer are passed into a cross-modality fusion module with the extracted textual embedding to calculate cross-modality self-attention. Then, we concatenate all updated visual or textual features from different levels respectively, and utilize a fully connected layer to map them into the original dimension. Finally, all features are concatenated and fed into a regression head to predict referred object regions. The regression head composes of three fully connected layers.
\vspace{5pt}
\section{Experiments}
\label{experiments}
\textbf{Dataset and setups.} Following previous visual grounding methods~\cite{deng2021transvg, yang2020improving}, we evaluate our method on five datasets: RefCOCO~\cite{yu2016modeling}, RefCOCO+~\cite{yu2016modeling}, RefCOCOg~\cite{mao2016generation}, ReferItGame~\cite{kazemzadeh2014referitgame}, and Flickr30K Entities~\cite{plummer2015flickr30k}. We follow the same train/val/test splits from~\cite{deng2021transvg} for all datasets. The number of training images in these five datasets are 16994, 16992, 24698, 8994, and 29779. Note that we don't use any manual annotations during the training stage, they are only leveraged for evaluation.

\begin{table*}[t]\footnotesize
\caption{Comparison with state-of-the-art methods on RefCOCO~\cite{yu2016modeling}, RefCOCO+~\cite{yu2016modeling} and RefCOCOg~\cite{mao2016generation} datasets in terms of top-1 accuracy (\%). \textit{``Sup."} refers to supervision level: No (without annotation), Weak (only annotated queries), Full (annotated bbox-query pairs). The best two results with supervision level of No and Weak are \textbf{bold-faced} and \underline{underlined}, respectively.}
\vspace{-12pt}
\begin{center}
\resizebox{2.1\columnwidth}{!}{%
\begin{tabular}{p{1.8cm} | p{1.6cm}<{\centering} | p{0.7cm}<{\centering} | p{0.8cm}<{\centering} p{0.8cm}<{\centering} p{0.8cm}<{\centering} | p{0.8cm}<{\centering} p{0.8cm}<{\centering} p{0.8cm}<{\centering} | p{0.8cm}<{\centering} p{0.8cm}<{\centering} p{0.8cm}<{\centering}}
    \toprule
\multirow{2}{*}{Method} & \multirow{2}{*}{Published on} & \multirow{2}{*}{Sup.} & \multicolumn{3}{c|}{RefCOCO} & \multicolumn{3}{c|}{RefCOCO+} & \multicolumn{3}{c}{RefCOCOg} \\
 & & & val & testA & testB & val & testA & testB & val-g & val-u & test-u \\
    \midrule
    \midrule
CPT \cite{yao2021cpt} & \textit{arXiv'21} & \multirow{2}{*}{No} & 32.20 & 36.10 & 30.30 & 31.90 & 35.20 & 28.80 & - & \underline{36.70} & \underline{36.50} \\
\textbf{Ours} & \textit{CVPR'22} & & \textbf{56.02} & \textbf{58.25} & \textbf{54.13} & \underline{38.88} & \textbf{45.06} & 32.13 & \textbf{49.82} & \textbf{46.25} & \textbf{47.44} \\
    \midrule
VC \cite{zhang2018grounding} & \textit{CVPR'18} & \multirow{4}{*}{Weak} & - & 33.29 & 30.13 & - & 34.60 & 31.58 & 33.79 & - & - \\
ARN \cite{liu2019adaptive} & \textit{ICCV'19} & & 34.26 & 36.43 & 33.07 & 34.53 & 36.01 & 33.75 & 33.75 & - & - \\
KPRN \cite{liu2019knowledge} & \textit{ACMMM'19} & & 35.04 & 34.74 & 36.98 & 35.96 & 35.24 & \underline{36.96} & 33.56 & - & - \\
DTWREG \cite{sun2021discriminative} & \textit{TPAMI'21} & & \underline{39.21} & \underline{41.14} & \underline{37.72} & \textbf{39.18} & \underline{40.10} & \textbf{38.08} & \underline{43.24} & - & - \\
    \midrule
MAttNet \cite{yu2018mattnet} & \textit{CVPR'18} & \multirow{5}{*}{Full} & 76.65 & 81.14 & 69.99 & 65.33 & 71.62 & 56.02 & - & 66.58 & 67.27 \\
NMTree \cite{liu2019learning} & \textit{ICCV'19} & & 76.41 & 81.21 & 70.09 & 66.46 & 72.02 & 57.52 & 64.62 & 65.87 & 66.44 \\
FAOA \cite{yang2019fast} & \textit{ICCV'19} & & 72.54 & 74.35 & 68.50 & 56.81 & 60.23 & 49.60 & 56.12 & 61.33 & 60.36 \\
ReSC \cite{yang2020improving} & \textit{ECCV'20} & & 77.63 & 80.45 & 72.30 & 63.59 & 68.36 & 56.81 & 63.12 & 67.30 & 67.20 \\
TransVG \cite{deng2021transvg} & \textit{ICCV'21} & & 80.32 & 82.67 & 78.12 & 63.50 & 68.15 & 55.63 & 66.56 & 67.66 & 67.44 \\
    \bottomrule
\end{tabular}%
}
\end{center}
\label{tab:SOTA1}
\vspace{-15pt}	
\end{table*}

\textbf{Implementation details.} We choose a pre-trained detector~\cite{anderson2018bottom} and attribute classifier~\cite{anderson2018bottom} on Visual Genome dataset~\cite{krishna2017visual}, which contains 1600 object and 400 attribute categories. As we mentioned in Sec.~\ref{pseudo}, we select top-$N$ and sample up to $M$ pseudo-queries for each image. Specifically, on RefCOCO, we select top-3 objects according to the detection confidence and uniformly sample 6 pseudo-queries from all possible candidates. As for RefCOCO+, RefCOCOg, ReferItGame, and Flickr30K Entities, we use top-3 objects/12 pseudo-queries, top-2 objects/4 pseudo-queries, top-6 objects/15 pseudo-queries, and top-7 objects/28 pseudo-queries, respectively.

\textbf{Training details.}
All our experiments are conducted under Pytorch framework~\cite{paszke2019pytorch} with 8 RTX3090 GPUs. Our visual-language model is end-to-end optimized with AdamW. The initial learning rate is set to $2.5\times10^{-5}$ for the visual and language encoder and $2.5\times10^{-4}$ for the cross-modality fusion module. The batch size is 256. All the datasets use cosine learning rate schedule except Flickr30K Entities which adopts exponential decay schedule with 0.85 decay rate. Our model is trained with 10 epochs on RefCOCO, RefCOCOg, and ReferItGame, 20 epochs on RefCOCO+ and Flickr30K Entities. The data augmentations that we utilize are following TransVG~\cite{deng2021transvg}, \textit{e.g.,} RandomResizeCrop, RandomHorizontalFlip and ColorJitter.

\subsection{Comparison with State-of-the-art Methods}
We report comparison results with existing unsupervised~\cite{yao2021cpt, wang2019phrase,yeh2018unsupervised} and weakly-supervised~\cite{sun2021discriminative,wang2021improving,liu2021relation} methods. Note that the weakly-supervised methods are trained with expensive annotated queries. As references, the performance of fully-supervised~\cite{deng2021transvg,yang2020improving} methods are showed as an upper bound. Specifically, we show the top-1 accuracy (\%) results following previous works~\cite{wang2021improving, liu2021relation}. The predicted bounding boxes are regarded as correct if the Jaccard overlaps between them and the ground truth are above 0.5.

\begin{table}[t]\footnotesize
\caption{Comparison with state-of-the-art methods on ReferItGame~\cite{kazemzadeh2014referitgame} and Flickr30K Entities~\cite{plummer2015flickr30k} in terms of top-1 accuracy (\%). \textit{``Sup."} refers to supervision level: No (without annotation), Weak (only annotated queries), Full (annotated bbox-query pairs). The best two results with supervision level of No and Weak are \textbf{bold-faced} and \underline{underlined}, respectively.}
\vspace{-12pt}
\begin{center}
\resizebox{\columnwidth}{!}{%
\begin{tabular}{p{2.2cm} | p{1.5cm}<{\centering} | p{0.6cm}<{\centering} | p{1cm}<{\centering} | p{1cm}<{\centering}}
    \toprule
Method & Published on & Sup. & ReferIt & Flickr30K \\
    \midrule
    \midrule
Yeh \textit{et al}.~\cite{yeh2018unsupervised} & \textit{CVPR'18} & \multirow{3}{*}{No} & 36.93 & 20.91 \\
Wang \textit{et al}.~\cite{wang2019phrase} & \textit{ICCV'19} & & 26.48 & 50.49 \\
\textbf{Ours} & \textit{CVPR'22} & & \textbf{43.32} & \textbf{60.41}  \\
    \midrule
Chen \textit{et al}.~\cite{chen2018knowledge} & \textit{CVPR'18} & \multirow{6}{*}{Weak} & 33.67 & 46.61 \\
Zhao \textit{et al}.~\cite{zhao2018weakly} & \textit{CVPR'18} & & 33.10 & 13.61 \\
Liu \textit{et al}.~\cite{liu2019adaptive} & \textit{ICCV'19} & & 26.19 & - \\
Gupta \textit{et al}.~\cite{gupta2020contrastive} & \textit{ECCV'20} & & - & 51.67  \\
Liu \textit{et al}.~\cite{liu2021relation} & \textit{CVPR'21} & & 37.68 & \underline{59.27}  \\
Wang \textit{et al}.~\cite{wang2021improving} & \textit{CVPR'21} & & \underline{38.39} & 53.10  \\
    \midrule
Kovvuri \textit{et al}.~\cite{kovvuri2018pirc} & \textit{ACCV'18} & \multirow{5}{*}{Full} & 59.13 & 72.83 \\
Yu \textit{et al}.~\cite{yu2018rethinking} & \textit{IJCAI'18} & & 63.00 & 73.30 \\
Yang \textit{et al}.~\cite{yang2019fast} & \textit{ICCV'19} & & 60.67 & 68.71 \\
Yang \textit{et al}.~\cite{yang2020improving} & \textit{ECCV'20} & & 64.60 & 69.28 \\
Deng \textit{et al}.~\cite{deng2021transvg} & \textit{ICCV'21} & & 69.76 & 78.47 \\
    \bottomrule
\end{tabular}%
}
\end{center}
\label{tab:SOTA2}
\vspace{-15pt}	
\end{table}

\textbf{RefCOCO/RefCOCO+/RefCOCOg.} Our method's performances on RefCOCO, RefCOCO+ and RefCOCOg datasets are reported in Table~\ref{tab:SOTA1}. We compare our method with the existing state-of-the-art unsupervised method CPT~\cite{yao2021cpt} and weakly-supervised method DTWREG~\cite{sun2021discriminative}. Our method can easily surpass CPT by a remarkable margin on all three datasets (\textit{e.g.}, 23.82\%/22.15\%/23.83\% performance improvement on RefCOCO's val/testA/testB split respectively). When compared with the DTWREG method, our method can achieve better performances on RefCOCO and RefCOCOg datasets. Meanwhile, it can obtain comparable and superior performances on val and testA split of RefCOCO+ dataset. Although we can see that there's an accuracy gap compared with DTWREG on testB split, our method still gets a large performance gain over CPT. Note that without leveraging any manually labeled queries of RefCOCO+'s training split, our method can still reach considerable performance. All the experiments demonstrate that our generated pseudo-queries can provide effective supervision signals for visual grounding task.


\textbf{ReferItGame.} In Table~\ref{tab:SOTA2}, we show the comparisons with other existing visual grounding methods on ReferItGame dataset. Our method can achieve 43.32\% top-1 accuracy, which outperforms all unsupervised and weakly-supervised methods. Especially, compared with the state-of-the-art weakly-supervised method~\cite{wang2021improving}, which can achieve 38.39\% top-1 accuracy, our method can obtain 4.93\% performance improvement without using any annotated labels. These results can demonstrate the superiority of our proposed method.

\textbf{Flickr30K Entities.} The results on Flickr30K Entities dataset is shown in Table~\ref{tab:SOTA2}. It can be observed that our method can still achieve surprising 60.41\% top-1 accuracy which outperforms the state-of-the-art weakly-supervised method~\cite{liu2021relation} by 1.14\%. Considering the scale of Flick30K Entities which consists of 427K manually annotated referred expressions, our method still achieves remarkable performance without any training label. As for other methods without using manual labels, our method can easily surpass \cite{yeh2018unsupervised} and \cite{wang2019phrase} with 39.50\% and 9.92\% absolute performance improvement, respectively.

\begin{figure}[t]
\centering
\includegraphics[scale=0.55]{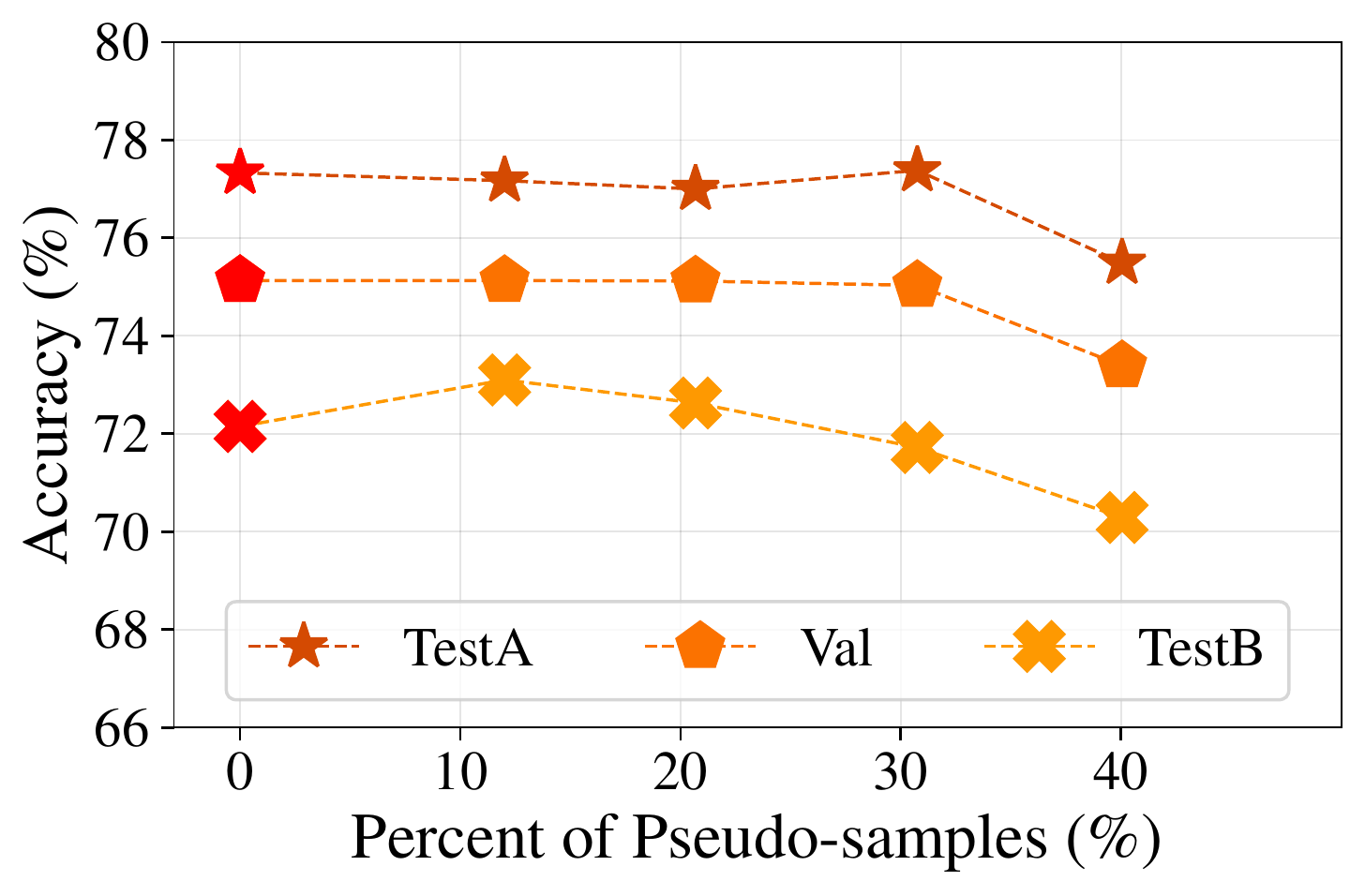}
\caption{Experiments of reducing the manual labeling cost on RefCOCO~\cite{yu2016modeling}. We replace the manual labels whose queries contain spatial relationships with our pseudo-samples.}
\label{fig:reduce_cost}
\centering
\vspace{-12pt}
\end{figure}

\textbf{Explanations of the gain over weakly-supervised methods.} First, the core of visual grounding task is learning the \textbf{\textit{correspondence}} between visual and linguistic modalities which relies heavily on the correct mapping between image regions and queries inside training data. A \textbf{\textit{key difference}} between our approach and weakly-supervised methods is that we can generate corresponding queries for the detected object which guarantees the \textbf{\textit{correctness of mapping}} between two modalities. Although weakly-supervised methods have annotated queries, they lack key supervision signals that are the region-level correspondence between two modalities. Second, our model jointly optimizes the features from two modalities which allows the model to learn a better correspondence while weakly-supervised methods~\cite{liu2021relation, sun2021discriminative, wang2021improving} only update the language model leaving the visual model fixed. 
\label{comparison}

\begin{figure}[t]
    \centering
    \includegraphics[width=\columnwidth]{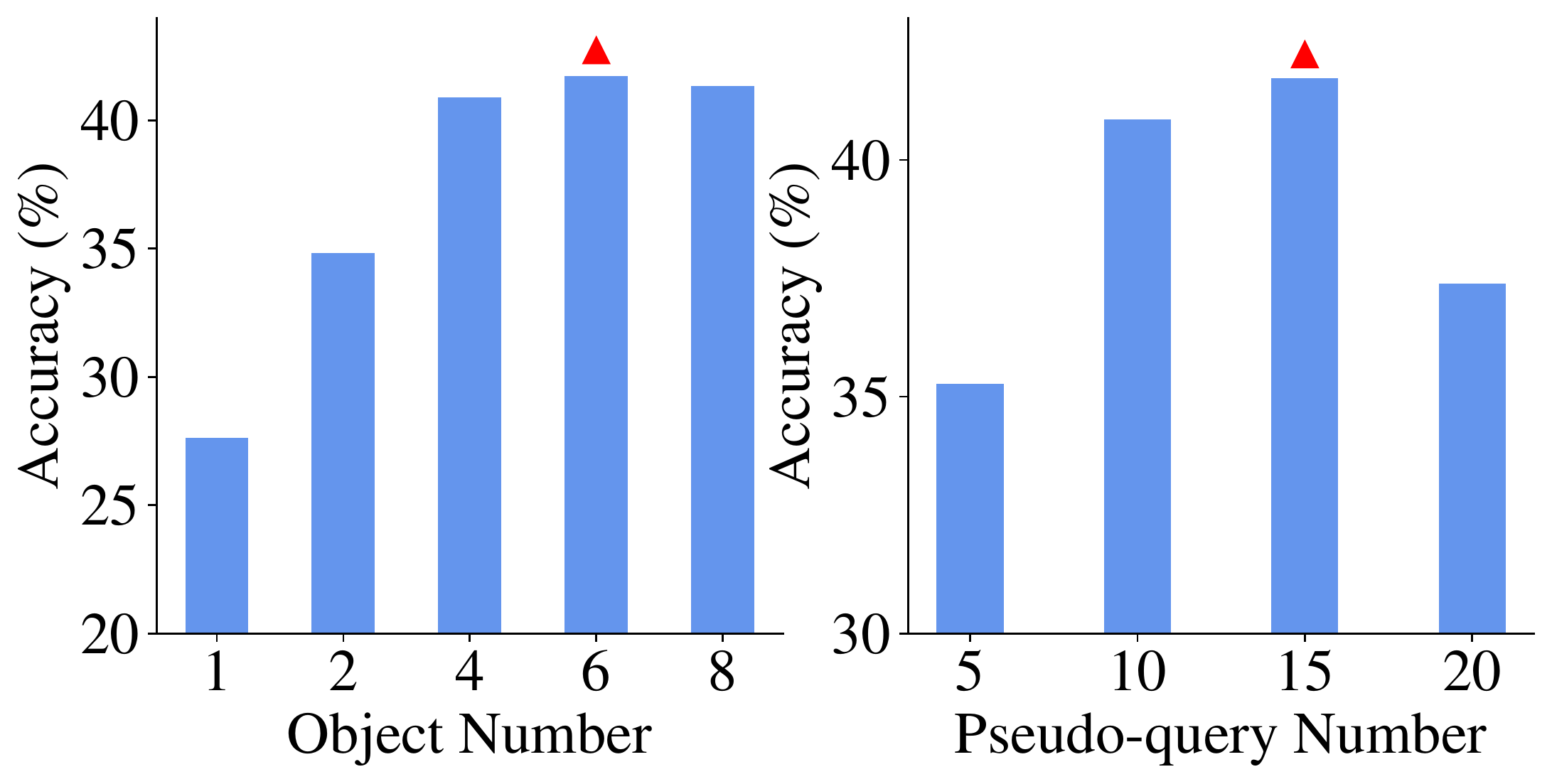}
    \vspace{-12pt}
    \caption{Left: Ablation of object number. Right: Ablation of pseudo-query number. Both are conducted on ReferItGame~\cite{mao2016generation}.}
    \label{fig:ablation_of_object_query}
    \vspace{-10pt}
\end{figure}

\subsection{Improving the Efficiency of Manual Labeling.}
In Figure~\ref{fig:reduce_cost}, we perform experiments with the same hyper-parameters as Sec.~\ref{experiments} on RefCOCO~\cite{yu2016modeling} to verify the effectiveness of our pseudo-samples, \textit{i.e.}, pseudo image region-query pairs, by replacing the manually annotated labels whose queries contain spatial relationships. The baseline is our model trained in a fully-supervised manner. Note that the query prompt module is not applied in this experiment. As we can see, compared to the fully-supervised setting, substituting 12.01$\%$, 20.68$\%$, and 30.75$\%$ manually annotated labels with our generated pseudo region-query pairs do not degrade the original performance. In such a situation, about \textbf{31\%} of human annotation costs can be reduced. Consequently, our method can be utilized to automatically annotate one of the dominant components, \textit{i.e.} spatial relationship, in language queries, which significantly improves the efficiency of manual labeling.

\vspace{5pt}
\subsection{Ablation Study}
\label{ablation}
In this section, we conduct extensive ablation experiments to demonstrate the effectiveness of each proposed component and the rationality of hyper-parameters setting. Most of the following experiments are conducted on ReferItGame~\cite{kazemzadeh2014referitgame} dataset and we report the top-1 accuracy. The model is trained with the same hyper-parameters as Sec.~\ref{experiments}.

\textbf{Number of nouns.} We investigate the impact of utilizing different number of nouns (objects) in Figure~\ref{fig:ablation_of_object_query}(a). Increasing the number of nouns can produce more pseudo samples which boosts the performance of our model, as shown in Figure~\ref{fig:ablation_of_object_query}(a). In our experiments, we use the detection confidence as a criterion to select salient objects. If the number of nouns is too large, the likelihood of detecting low confidence objects which are inconspicuous will grow. We empirically find that, on ReferItGame dataset, the model reaches its peak performance when the number of nouns is 6. Once the number of nouns exceeds 6, the performance starts to degrade. Thus, we use the top-6 object proposals on the ReferItGame dataset.

\textbf{Number of pseudo-queries.} Another essential factor is the number of pseudo-queries. We study the influence of sampling different number of pseudo-queries in Figure~\ref{fig:ablation_of_object_query}(b). The candidates of pseudo-queries are generated following templates in Appendix. As shown in Figure~\ref{fig:ablation_of_object_query}(b), our model performs best when the sampling number of pseudo-queries is 15. If the sampling number is too small, we will miss plenty of useful candidates which hinders the improvement of model performance. Meanwhile, note that not every candidate provides the correct supervision signal. Thus, overly sampling candidates will also hurt the performance. Finally, we sample up to 15 pseudo-queries.

\textbf{Effectiveness of integrating attributes.} We empirically support the effectiveness of introducing attributes into pseudo-queries by comparing them with those lacking attributes. As shown in Table \ref{tab:ablation_each_components}, generating pseudo-queries with nouns and attributes clearly surpasses the one that only has nouns on RefCOCO and ReferItGame. Moreover, adding attributes into pseudo-queries with nouns and relationships can further boost the performance. Thus, we demonstrate that incorporating the attribute into pseudo-queries helps models to comprehend the scenes better.

\textbf{Effectiveness of generating relationships.} As we mentioned in Sec.~\ref{pseudo}, spatial relationship is the most essential component. With only nouns, models are still far from comprehensively understanding scenes. The ablation study of generating relationships that supports our proposition is reported in Table~\ref{tab:ablation_each_components}. Compared with pseudo-queries with only nouns, generating relationships with our method outperforms it overwhelmingly by $\textbf{26.67\%}$ on RefCOCO. In sum, experimental results show that incorporating spatial relationships into pseudo-queries can significantly enhance the model's capability of understanding scenes.

\begin{table}[t]\footnotesize
\caption{Ablations of each component on RefCOCO~\cite{yu2016modeling} and ReferItGame~\cite{mao2016generation}. \textit{``Attr"} and \textit{``Rela"} denote attribute and relationship, respectively. \textit{``Prompt"} represents the query prompt module. \textit{``ML-CMA"} means the proposed multi-level cross-modality attention.}
\vspace{-12pt}
\begin{center}
\resizebox{\columnwidth}{!}{
\begin{tabular}{p{0.6cm}<{\centering} p{0.6cm}<{\centering} p{0.6cm}<{\centering} p{1.2cm}<{\centering} p{0.8cm}<{\centering} | p{1.8cm}<{\centering} p{1.8cm}<{\centering}}
    \toprule
Noun & Attr & Rela & ML-CMA & Prompt & RefCOCO & ReferIt \\
    \midrule
    \midrule
\cmark & & & & & 22.04 & 27.91 \\
\cmark & \cmark & & & & 31.30 \textcolor{blue} {($\uparrow$9.26)} & 31.33 \textcolor{blue} {($\uparrow$3.42)} \\
\cmark & & \cmark & & & 48.71 \textcolor{blue} {($\uparrow$\textbf{26.67})} & 39.26 \textcolor{blue} {($\uparrow$\textbf{11.35})} \\
\cmark & \cmark & \cmark & & & 53.39 \textcolor{blue} {($\uparrow$31.35)} & 40.37 \textcolor{blue} {($\uparrow$12.46)} \\
\midrule
\cmark & \cmark & \cmark & \cmark & & 55.16 \textcolor{blue} {($\uparrow$1.77)} & 41.72 \textcolor{blue}{($\uparrow$1.35)} \\ 
\cmark & \cmark & \cmark & \cmark & \cmark & 56.02 \textcolor{blue} {($\uparrow$2.63)} & 43.32 \textcolor{blue}{($\uparrow$2.95)} \\ 
    \bottomrule
\end{tabular}
}
\end{center}
\label{tab:ablation_each_components}
\vspace{-15pt}	
\end{table}

\begin{figure*}[t]
\centering
\includegraphics[scale=0.47]{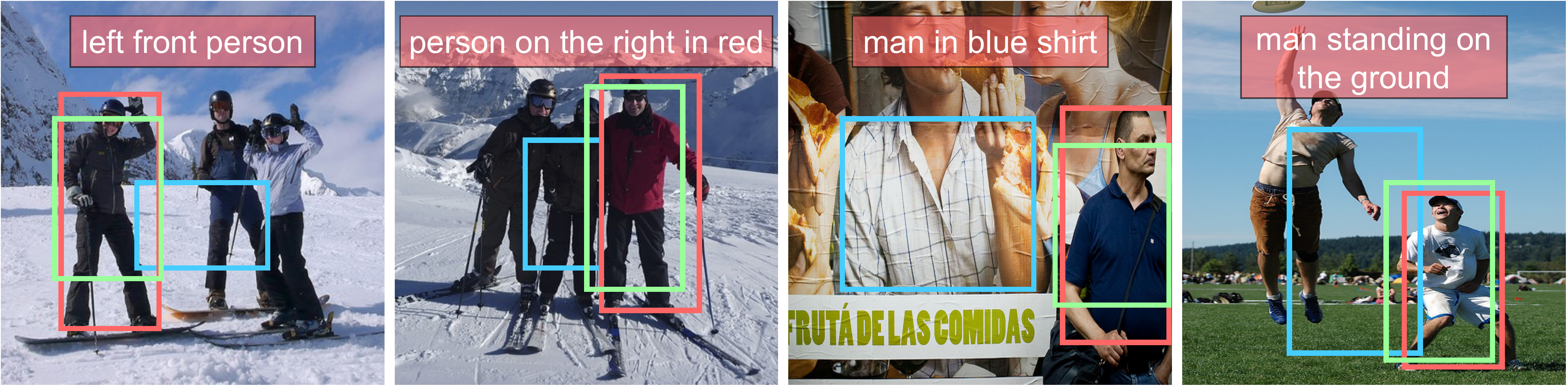}
\caption{Four visualization examples of detection results. The \textcolor{myred}{red bounding boxes} and queries are ground truth. The \textcolor{mygreen}{green bounding boxes} are detected by the model that trained on pseudo-samples generated with nouns, attributes, and relationships. The \textcolor{myblue}{blue bounding boxes} are detected by the model that trained on pseudo-samples generated with only nouns.}
\label{fig:vis_detection_result}
\centering
\end{figure*}

\begin{figure*}[t]
\centering
\includegraphics[scale=0.47]{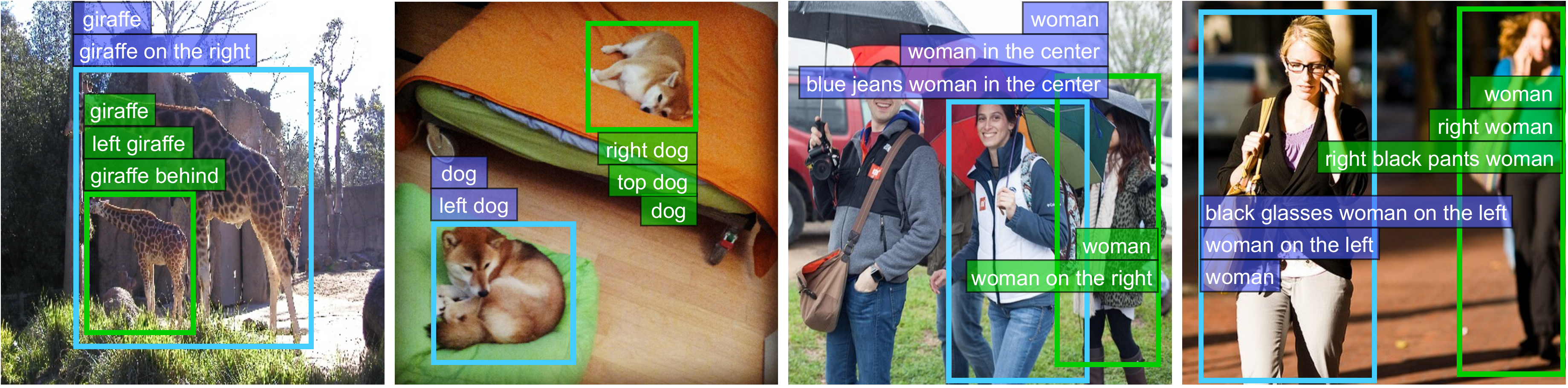}
\caption{Four visualization examples of pseudo-sample generated by our method. We use blue and green to distinguish two objects.}
\label{fig:vis_pseudo_sample}
\centering
\end{figure*}

\textbf{Effectiveness of query prompts.} In Table~\ref{tab:ablation_each_components}, we show that prompts help to excavate the hidden knowledge of the pre-trained language model, and consequently, boost the performance. On ReferItGame, the well-designed prompt \emph{``which region does the text \{pseudo-query\} describe?"} improves the performance by $1.60\%$. Meanwhile, on RefCOCO, the prompt \emph{``find the region that corresponds to the description \{pseudo-query\}"} improves the performance by $0.86\%$. On the other hand, we find that hand-designed prompts are not robust enough across all the datasets.

\textbf{Effectiveness of cross-modality fusion module.} We further investigate the contribution of the cross-modality fusion module in Table~\ref{tab:ablation_each_components}. On the basis of the pseudo-query generation module, the proposed cross-modality fusion module can further improve the performance by $1.77\%$ and $1.35\%$ on RefCOCO and ReferItGame, respectively. 

\vspace{5pt}
\subsection{Qualitative Analysis}

\label{analysis}
To further figure out the importance of spatial relationships and attributes, in Figure~\ref{fig:vis_detection_result}, we show four detection examples of our models trained on generated pseudo-queries with or without spatial relationships and attributes on RefCOCO dataset. In the first two examples, we can easily observe that the model trained with relationships locates target objects much better than the one trained without relationship component. In the last two example, the key factor to locate the queried man is leveraging the attributes ``\textit{blue}" and ``\textit{standing}". Obviously, with the above analysis, we can conclude that the proposed spatial relationship and attribute play an essential role in accurately grounding referred objects with given language queries. In addition, we also display four generated pseudo region-query pairs on RefCOCO dataset in  Figure~\ref{fig:vis_pseudo_sample}.

\subsection{Limitation}
Although our method achieves superior performances on five datasets, there are still two limitations. First, when it comes to pseudo language queries, there may be some incorrect queries which harms final performance. Second, the generated pseudo-queries are simple, other relationships can be explored in the future.
\section{Conclusion}
In this paper, we make the first attempt to introduce a pseudo-query based visual grounding method called Pseudo-Q. Firstly, we propose a pseudo-query generation module to automatically produce pseudo region-query pairs for supervised training. Then, we present a query prompt module, so that generated pseudo language queries can be tailored specifically for visual grounding task. Finally, in order to sufficiently model the relationships between visual regions and language-queries, we develop a visual-language model equipped with multi-level cross-modality attention. Extensive experiments have shown that our method can not only achieve superior performances on five datasets, but also dramatically reduce manual labeling costs.
\section*{Acknowledgement}
This work is supported in part by the National Science and Technology Major Project of the Ministry of Science and Technology of China under Grants 2018AAA0100701, the National Natural Science Foundation of China under Grants 61906106 and 62022048, the Guoqiang Institute of Tsinghua University and Beijing Academy of Artificial Intelligence.

{\small
\bibliographystyle{ieee_fullname}
\bibliography{egbib}
}

\appendix
\section*{Appendix}
\addcontentsline{toc}{section}{Appendices}
\renewcommand{\thesubsection}{\Alph{subsection}}
\section{Statistics of RefCOCO Dataset}
\label{sec:statistic}
\begin{figure}[h]
    \centering
    \subfloat[]{
        \includegraphics[width=0.47\columnwidth]{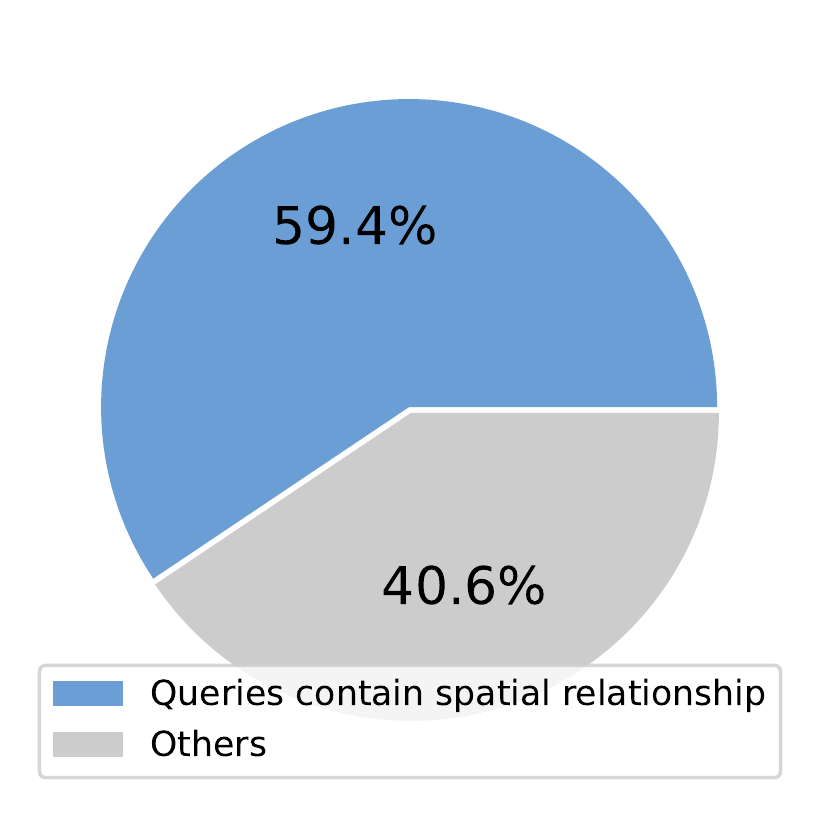}
    }
    \subfloat[]{
	\includegraphics[width=0.47\columnwidth]{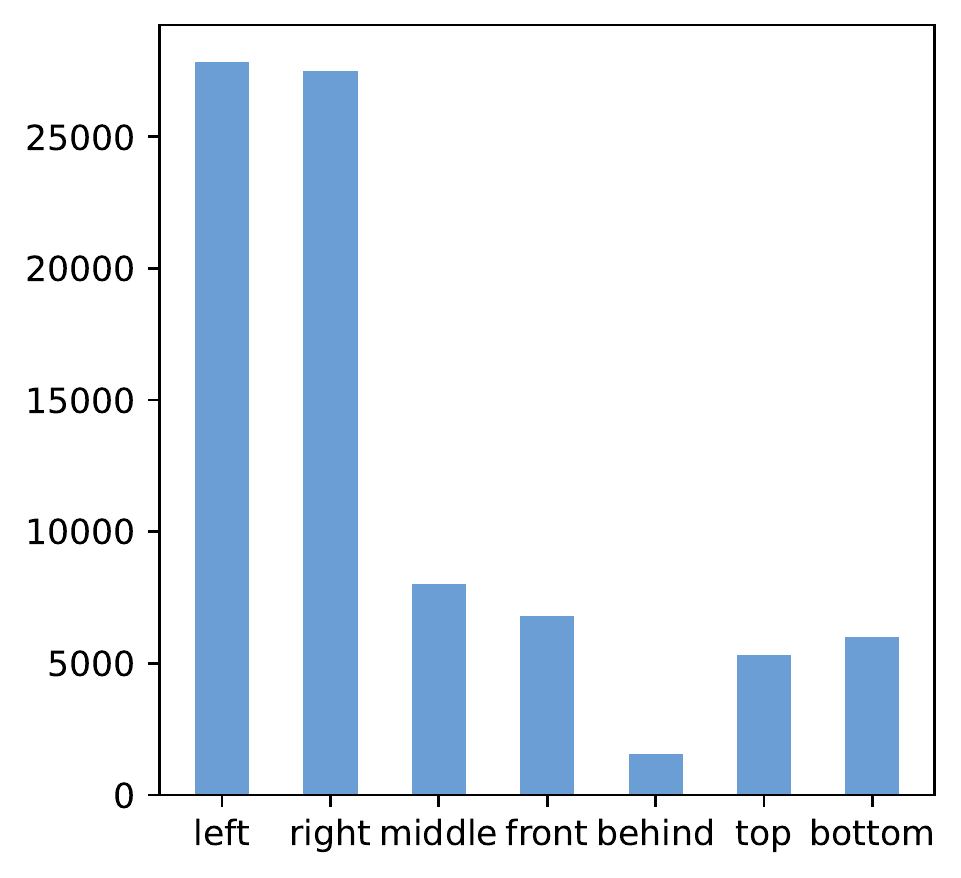}
    }
    \caption{\textbf{Statistics of the training set of RefCOCO~\cite{yu2016modeling} dataset.} (a): The percent of language queries that contain spatial relationships. (b): The number of different spatial relationships.}
    \label{fig:statistics}
    \vspace{-15pt}
\end{figure}

In Figure~\ref{fig:statistics}, we show the statistics of the training set of RefCOCO~\cite{yu2016modeling} dataset to demonstrate spatial relationship is one of the dominant components in language queries. As we can see, spatial relationships exists in almost $60\%$ of queries. Furthermore, the most common spatial relationships in RefCOCO are \emph{left} and \emph{right}. In addition, other spatial relationships, \textit{i.e.,} \emph{middle}, \emph{front}, \emph{top}, and \emph{bottom}, are also frequently found in language queries.

\section{Pseudo-Query Templates}
Our pseuod-queries are generated following the templates shown in Table.~\ref{tab:pseudo_query_templates}. All the possible templates is considered in our method for the purpose of obtaining as many candidate pseudo-samples as possible. Honestly, this strategy will inevitably produce some ungrammatical pseudo-samples. Our approach is similar to all the pseudo-label based methods, such as semi-supervised learning, which can’t guarantee every single pseudo-query is correct. Overall, these pseudo-queries provide valuable supervision signals and eventually benefit the training of the model.

\begin{table}[h]
\caption{Pseudo-query templates. \emph{Attr} and \emph{Rela} represents attribute and relationship, respectively.}
\vspace{-10pt}
\begin{center}
\resizebox{\columnwidth}{!}{
\begin{tabular}{p{3.6cm}<{\centering} | p{5cm}<{\centering}}
    \toprule
Pseudo Query Template & Example  \\
    \midrule
    \midrule
\textit{\{Noun\}} & ``man",  ``building" etc. \\
    \midrule
\textit{\{Noun\}} \textit{\{Attr\}} & ``man standing" etc. \\
\textit{\{Attr\}} \textit{\{Noun\}} & ``talk man", ``wooden building" etc. \\ 
    \midrule
\textit{\{Noun\}} \textit{\{Rela\}} & ``man on the right" etc. \\
\textit{\{Rela\}} \textit{\{Noun\}} & ``center man", ``left building" etc. \\ 
    \midrule
\textit{\{Noun\}} \textit{\{Attr\}} \textit{\{Rela\}} & ``man standing on the right" etc. \\
\textit{\{Noun\}} \textit{\{Rela\}} \textit{\{Attr\}} & ``man right standing" etc. \\
\textit{\{Attr\}} \textit{\{Noun\}} \textit{\{Rela\}} & ``standing man on the right" etc. \\
\textit{\{Attr\}} \textit{\{Rela\}} \textit{\{Noun\}} & ``standing right man" etc. \\
\textit{\{Rela\}} \textit{\{Noun\}} \textit{\{Attr\}} & ``right man standing" etc. \\
\textit{\{Rela\}} \textit{\{Attr\}} \textit{\{Noun\}} & ``right standing man" etc. \\
    \bottomrule
\end{tabular}
}
\end{center}
\label{tab:pseudo_query_templates}
\vspace{-15pt}	
\end{table}

\section{Visual-Language Model}
\label{sec:intro}
In this section, we provide more details about the architecture of the visual encoder and the language encoder.

In the visual encoder, a CNN backbone and a transformer-based network are stacked sequentially for image feature extraction. The CNN backbone is a ResNet-50 model~\cite{he2016deep} pre-trained on ImageNet~\cite{deng2009imagenet}, and the transformer-based network is the encoder part of DETR network~\cite{carion2020end} which consists of six transformer layers. Moreover, the pre-trained weights of DETR are utilized for initialization. The output feature maps of the ResNet-50 are fed into a 1 $\times$ 1 convolutional layer for dimension reduction. Then, they are flattened into 1D vectors for the transformer network.

In the language encoder, a token embedding layer and a linguistic transformer are employed to extract textual features. Specifically, the token embedding layer is leveraged to convert the discrete words into continuous language vectors. Since BERT~\cite{devlin2018bert} has been successfully applied for text feature extraction, the BERT architecture which has 12 transformer layers is adopted as the linguistic transformer.

\end{document}